\documentclass[11pt]{article}
\usepackage[utf8]{inputenc}

\usepackage{xcolor}
\usepackage{natbib}
\usepackage{cleveref}
\usepackage{footnote}

\usepackage{amsfonts,url,epsfig}
\usepackage{geometry} 
\usepackage{sectsty} 
\usepackage[normalem]{ulem}

\usepackage{nohyperref}
\usepackage{todonotes}
\makeatletter
\renewcommand\@biblabel[1]{#1.}
\makeatother

\sectionfont{\sffamily\bfseries\upshape\large}
\subsectionfont{\sffamily\bfseries\upshape\normalsize} 
\subsubsectionfont{\sffamily\mdseries\upshape\normalsize}
\makeatletter
\renewcommand\@seccntformat[1]{\csname the#1\endcsname.\quad}

\def\@maketitle{%
  \begin{center}%
  \let \footnote \thanks
    {\large \@title \par}%
    {\normalsize
      \begin{tabular}[t]{c}%
        \@author
      \end{tabular}\par}%
    {\small \@date}%
  \end{center}%
}
\makeatother

\title{\bf Toward a Taxonomy of Trust for Probabilistic Machine Learning\footnote{We thank the U.S. National Science Foundation, Office of Naval Research, Institute for Education Sciences, National Institutes of Health, and Sloan Foundation for partial support of this work.}\vspace{.1in}}
\author{Tamara Broderick$^1$, Andrew Gelman$^{2,3}$, Rachael Meager$^{4}$, Anna L. Smith$^5$, and Tian Zheng$^2$}
\date{%
\vspace{.1in}
$^1$Department of Electrical Engineering and  Computer Science, Massachusetts
Institute of Technology;\\%
$^2$Department of Statistics, Columbia University;\\%
$^3$Department of Political Science, Columbia University;\\%
$^4$Department of Economics, The London School of Economics and Political Science;\\%
$^5$Department of Statistics, University of Kentucky.\\%
}

\begin{document}

\maketitle

\begin{abstract}
Probabilistic machine learning increasingly informs critical decisions in medicine, economics, politics, and beyond. We need evidence to support that the resulting decisions are well-founded. To aid development of trust in these decisions, we develop a taxonomy delineating where trust in an analysis can break down: (1) in the translation of real-world goals to goals on a particular set of available training data, (2) in the translation of abstract goals on the training data to a concrete mathematical problem, (3) in the use of an algorithm to solve the stated mathematical problem, and (4) in the use of a particular code implementation of the chosen algorithm. We detail how trust can fail at each step and illustrate our taxonomy with two case studies: an analysis of the efficacy of microcredit and The Economist's predictions of the 2020 US presidential election. Finally, we describe a wide variety of methods that can be used to increase trust at each step of our taxonomy. The use of our taxonomy highlights steps where existing research work on trust tends to concentrate and also steps where establishing trust is particularly challenging.
\end{abstract}

\noindent
``Science, at its core, is a social phenomenon. It is a reflection of people, of our relationships, and of our institutions. When we provide inputs to the algorithm, when we program the device, when we design, test, and research, we are making human choices---choices that bring our social world to bear in a new and powerful way.'' --Alondra Nelson, Deputy Director for Science and Society, White House Office of Science and Technology Policy, 2021.

\section{Introduction}
Machine learning (ML) in general, and probabilistic methods in particular, are increasingly used to make major decisions in science, the social sciences, and engineering---with the potential to profoundly impact individuals' day-to-day lives. For instance, probabilistic methods have driven knowledge of the spread and effects of SARS-CoV-2 \citep{flaxman2020report,van2020aerosol,fischer2020assessment}, underlie election predictions at The Economist \citep{heidemanns_gelman_morris_2020, gelman_hullman_wlezien_etal_2020}, and can guide our understanding of the efficacy of microloans in alleviating poverty \citep{meager_2019}.
Given the large and growing impact of probabilistic machine learning (ML), it behooves us to make sure that its outputs are useful for its users' stated purposes.

There are many potential points, though, where a data analysis pipeline may break down. This issue becomes especially pressing as statistics and ML workflows become increasingly complex to face modern challenges. These challenges arise not just from the sheer size of the data but also from the inherent difficulty of the problems being studied. Big Data are messy data: confounded data rather than random samples, observational data rather than experiments, available data rather than direct measurements of underlying constructs of interest. To make relevant inferences from big data, we need to extrapolate from sample to population, from control to treatment group, and from measurements to latent variables. All of these steps require modeling. And Big Data needs Big Model: latent-variable models for psychological states or political ideologies, differential equation models in pharmacology, dynamic image analysis, and so forth. To meet modern challenges, we often fit models that are on the edge of our ability to compute and interact with in the real world. These models typically contain many assumptions and decision points. And we are often pushed to adjust for more factors to capture the complexity of our world. But models that adjust for lots of factors are hard to estimate. Fitting big models to big data requires scalability, so we must often turn to approximations such as Markov chain Monte Carlo, variational inference, and expectation propagation.

As this pipeline becomes more elaborate, there are more potential points of failure. But other complex endeavors succeed thanks in large part to extensive infrastructure: e.g., software engineering has testing and construction has scaffolding. It is similarly necessary to build an infrastructure to support the trustworthy creation and deployment of methods in probabilistic ML.
To this end, there exists a large literature addressing concerns of trust in data science---with key words including reproducibility, replicability, theoretical guarantees, stability, and interpretability.

To a new reader, it may not be obvious how these concerns relate to each other and what new work is needed in this area. In the present paper, we develop a \emph{taxonomy of trust}, splitting the probabilistic ML workflow (or data analysis pipeline, or inferential chain) into distinct constituent parts where trust can fail. Just as testing in software engineering benefits from modularity, we hope that the modularity of our taxonomy can facilitate more targeted work on trust concerns. We will use our taxonomy to highlight concerns that are relatively well-studied and those that could benefit from more attention.\footnote{We here focus on trust in the quality of decisions made using the results of a data analysis. By contrast, concerns about privacy and security are important and related to trust in the use of data and its artifacts, but these fall outside the scope of the present article.} 

We lay out the taxonomy in \Cref{sec:taxonomy_overview} and illustrate its parts with two case studies. 
We then use the taxonomy to organize and discuss different approaches to establishing trust in a data analysis---including existing work on reproducibility and replicability (\Cref{sec:establishing_trust}).

\section{Where trust can break down}
\label{sec:taxonomy_overview}

\begin{figure}[t]
\centerline{\includegraphics[width=\textwidth]{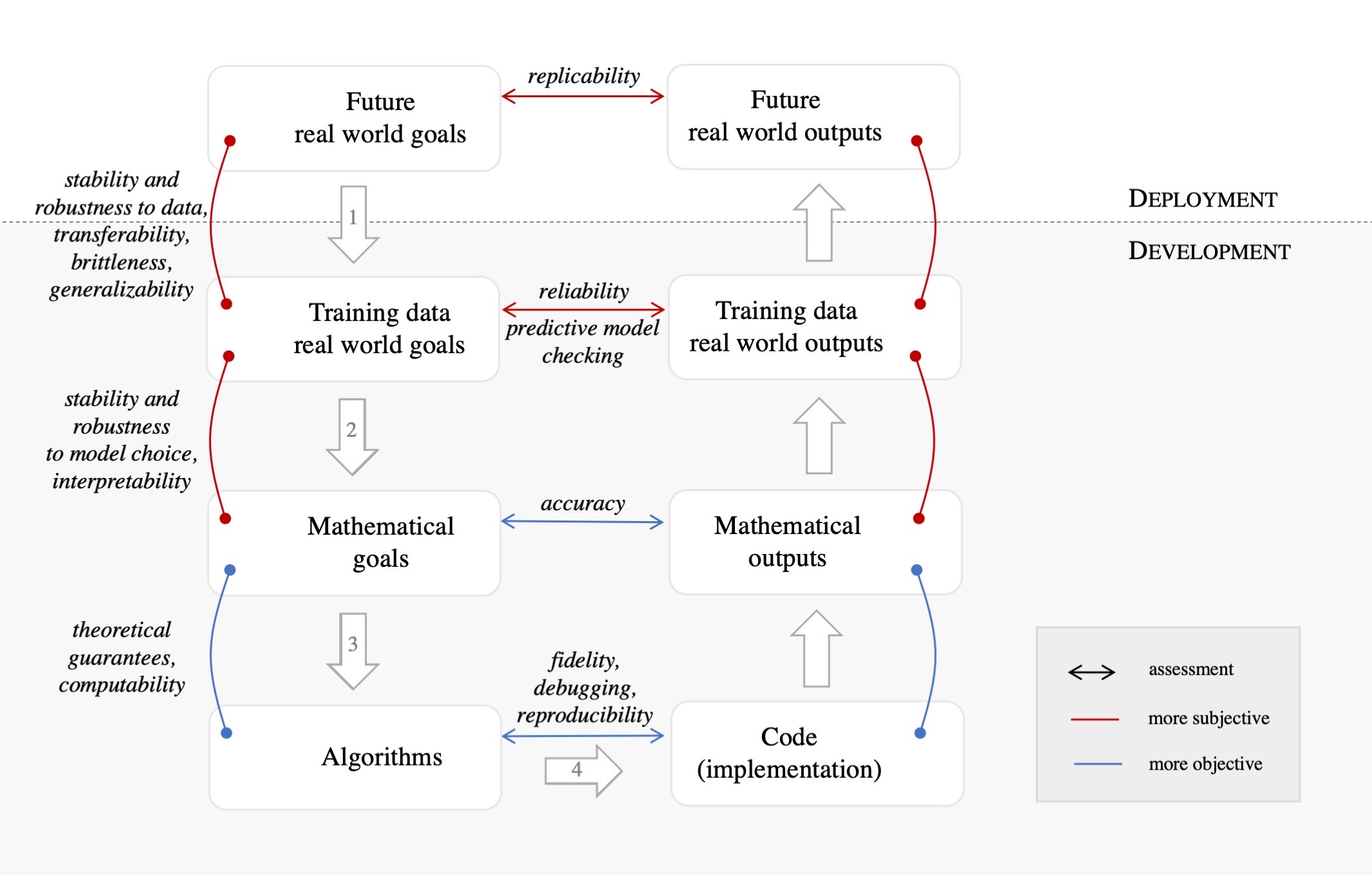}}
\caption{\em Diagram illustrating steps where trust can break down in a data analysis workflow. See \Cref{sec:taxonomy_overview} for a detailed description of the (non-italicized) steps. See \Cref{sec:establishing_trust} for a discussion of the italicized connectors, which include many of the characteristics of \citet{schwartz_down_jonas_etal_2021} for cultivating trust in AI systems.} 
\label{fig:trust}
\end{figure}

\Cref{fig:trust} gives a visualization of our \emph{taxonomy of trust}; this graphic provides a map of how probabilistic machine learning analyses interact with the real world. We first give a high-level overview of its steps, then describe how it shows where trust can break down in a data analysis, and finally give two case studies as examples.

We typically embark upon a data analysis because we seek to make an impact on our understanding of the world or generate future decision recommendations. From this perspective, an analysis is designed with ``Future real-world goals'' in mind (upper left of \Cref{fig:trust}). The data analyst then follows a rough workflow (on the left side of the figure) to turn this abstract goal into a concrete set of actions, which correspond to the numbered steps (arrows) in \Cref{fig:trust}.
\begin{enumerate}
    \item To serve these real-world goals, data must be gathered and processed for analysis. At this point, the analyst has reduced the problem to ``Training-data real-world goals.''
    \item The analyst chooses a model and expresses the real-world problem as a mathematical problem. The analyst has now reduced the problem to concrete ``Mathematical goals.'' 
    \item The analyst chooses particular ``Algorithms'' to solve the mathematical problem.
    \item The analyst runs the algorithm in practice using some particular ``Code (implementation).'' 
\end{enumerate}
We see the outputs of this workflow on the right side of \Cref{fig:trust}. The analyst runs their code and obtains a particular mathematical output. The output on the training data at hand is interpreted in the context of the real-world goal. Finally, the learned model is applied to future real-world outputs to make substantive conclusions and decisions.

Each numbered step in \Cref{fig:trust} in turn represents a point where trust can break down in a data analysis.
\begin{enumerate}
    \item Researchers typically face constraints in the cost of gathering, processing, or analyzing data, so the data set is necessarily a limited representation of the world. In choosing to approach our abstract question with a particular analysis of a particular set of data, we trust that the results from our data analysis are relevant to conclusions and decisions we will make at other places or times.
    \item Essentially all models are misspecified. In choosing to express our problem with a model and mathematical formalism, we are trusting that the model and formalism adequately capture the substantive goals of the analysis.
    \item Algorithms are often supported by theory, but based on assumptions that may not be perfectly achieved in practice. In choosing to solve our mathematical problem with a particular algorithm, we are trusting that the algorithm accomplishes the particular mathematical goal.
    \item Code is a precise way to manifest an algorithm, but it is difficult to avoid bugs and conceptual errors---especially in large and complex code bases. In choosing to execute our algorithm via code, we are trusting that the code is faithful to the algorithm.
\end{enumerate}

\subsection{Case study: Microcredit efficacy}
\label{sec:microcredit}

We use an example from applied economics that one of us has worked on \citep{meager_2019}, to demonstrate how the steps of \Cref{fig:trust} relate to trust in a data analysis.

\subsubsection*{Future real-world goals} 
Economists and policymakers would like to know if programs promoting microcredit (small loans in developing countries), have an overall beneficial effect---and, if they do, to use that information to inform decisions on subsidizing or disbursing microcredit across many potential locations. 
A related question of interest to economists is whether, in a given context, expanding access to the services offered by local microlenders would be beneficial for a local population.

\subsubsection*{Step 1 $\Rightarrow$ Training-data real-world goals} 
We analyzed data from seven different microfinance programs, each located in a different country. The countries range from Mexico to Mongolia, Bosnia to the Philippines, and the programs encompassed for-profit banks, government programs, and NGOs. Each individual dataset has been used to understand local efficacy of expanding access to microloans. In \citep{meager_2019}, we combined data across different settings and even somewhat different loan products to assess the geographically broader question of expanding microcredit in new locations.

In each of the individual studies, microcredit access was assigned in a randomized controlled trial (RCT) in order to prevent selection bias (on the part of the microlender or borrowers) from confounding the estimate of the effect of microcredit.

Moreover, researchers in each case decide what ``beneficial'' means, then what variables to measure to capture it, and further how to measure these variables using surveys or other tools. 
In the microcredit RCTs, the researchers all considered small-business profits as an outcome, but only five of the seven studies considered household consumption, and even fewer considered other components of well-being such as community health. All of these variables were carefully conceptualized and measured; e.g., based on contextual knowledge and previous research, someone decides whether, in rural Mexico, goats count as consumption or as investment. Some of the studies also recorded contextual covariates or characteristics of households to understand how they affect microcredit efficacy, but the relative importance of different factors in driving household outcomes (and thus likely moderating the effects of credit) such as consumption or business profits is itself an open research question.

\paragraph{Trust.}
Trust could break down in this step if the particular data analysis proved not to be useful for making future decisions. A researcher almost always makes choices that trade off the practicality and feasibility of collecting data and running the analysis against accurately capturing the state of the world. On the data collection side, there are an unending set of outcomes and characteristics of homes or small businesses that researchers could collect, but for time and funding reasons, the list must usually be limited.

The preference for randomized trials of microcredit reflects a desire to estimate unconfounded treatment effects even without being able to do any covariate adjustment, which partially addresses this concern, but might at times result in greater noise, or less representative sample sizes and thus greater extrapolation error.

Finally, the use of multiple studies provides hope that the results will generalize more than any single study, but full generalization relies on the assumption that the studies we have cover relevant aspects of a broader population of contexts in which we might make policy. And even if this assumption is satisfied, all studies occur in the past, but the policy decision gets made in the future; it is necessary to trust that the inference remains applicable under the new circumstances, or to make assumptions allowing appropriate adjustments to make this extrapolation.

\subsubsection*{Step 2 $\Rightarrow$ Mathematical goals.} 
The original RCT papers and the later meta-analysis all required a statistical model and choice of inference procedure. Economists often choose linear models for their interpretability. We chose a hierarchical linear model to allow partial pooling across the different studies, reflecting both their shared information and idiosyncrasies. Moreover, we took a Bayesian approach and provided a prior as part of the model.

We reported Bayesian posterior summaries such as posterior means and variances. Given a hierarchical model and non-conjugate priors, these must be approximated. So the mathematical goal is to report accurate approximations of these quantities for the particular chosen model.

\paragraph{Trust.} Trust could break down in this step since we know that models are essentially always misspecified in practice. Linear models of conditional mean dependence are highly interpretable but cannot capture nonlinear trends that may exist (and can be captured by other models). Even if the policy intervention is represented by a binary variable, linearity in raw household outcomes is incompatible with linearity in log space, and the distinction matters for extrapolating to new settings. Focusing on the mean itself is an influential choice: the mean effect of microcredit may be positive even if just a small portion of people actually benefit and even if some are harmed; in contrast, what we might really wish to understand is whether many people benefit from microcredit, or the community as a whole experiences net benefit. Moreover, reporting a posterior mean and variance alone, without additional visualizations, can hide posterior multimodality, heavy tails, or other distributional features that might cause us to pause and dig deeper into an analysis.

\subsubsection*{Step 3 $\Rightarrow$ Algorithms} 
We chose to approximate the Bayesian posterior mean and variance summaries using  Hamiltonian Monte Carlo (HMC) \citep{neal2011mcmc}.
In related work \citep{giordano2016fast}, we have considered variational Bayesian (VB) approximations of these posterior quantities instead; in the latter case, we employed mean-field VB approximations \citep{blei2017variational} as well as linear response corrections \citep{giordano2018covariances,giordano2015linear}.

\paragraph{Trust.} Trust could break down in this step if the algorithm did not in fact accomplish the mathematical goal. For instance, MCMC estimates of posterior expectations will converge to their exact values when an algorithm is run for enough time. But, despite a large literature on mixing rates, there are typically no specific guarantees on performance after running a particular MCMC algorithm for a particular finite time. Analogously,
VB will return a match to the exact posterior when used with a sufficiently large family of distributions. But in practice, the family of approximating distributions is limited, and the Kullback-Leibler (KL) divergence between the approximating and true posterior will be strictly greater than zero. Even moderate KL divergence values can correspond to arbitrarily large discrepancies between exact and reported posterior means and variances \citep{huggins2020validated}.

\subsubsection*{Step 4 $\Rightarrow$ Code} 
As increasingly complex methods are used for data analysis, a practitioner will often turn to existing software packages. Here we used Stan for HMC \citep{stan}. \citet{giordano2016fast} instead develops new code for VB.

\paragraph{Trust.} Trust could break down in this step if there are bugs in the code. There is robust software-engineering guidance on how to test code, including unit testing. Data analysts often rely on a software package for implementation and thus, to some extent, often outsource much (but certainly not all) of the establishment of trust in this step. Fitting the model in Stan, a well-tested probabilistic programming language, gave us confidence in the inferential part of our computation. But there is still the entire data and analysis pipeline to consider; recall the infamous Excel error of \citet{reinhart2010growth} \citep{herndon2014does,o2013forget}.

\subsection{Case study: Election forecasting}
\label{sec:election}

We similarly walk through the parts of our taxonomy, and how they relate to trust, with an example from our development of a public poll-aggregation algorithm for election forecasting.

\begin{figure}
\centerline{\includegraphics[width=.9\textwidth]{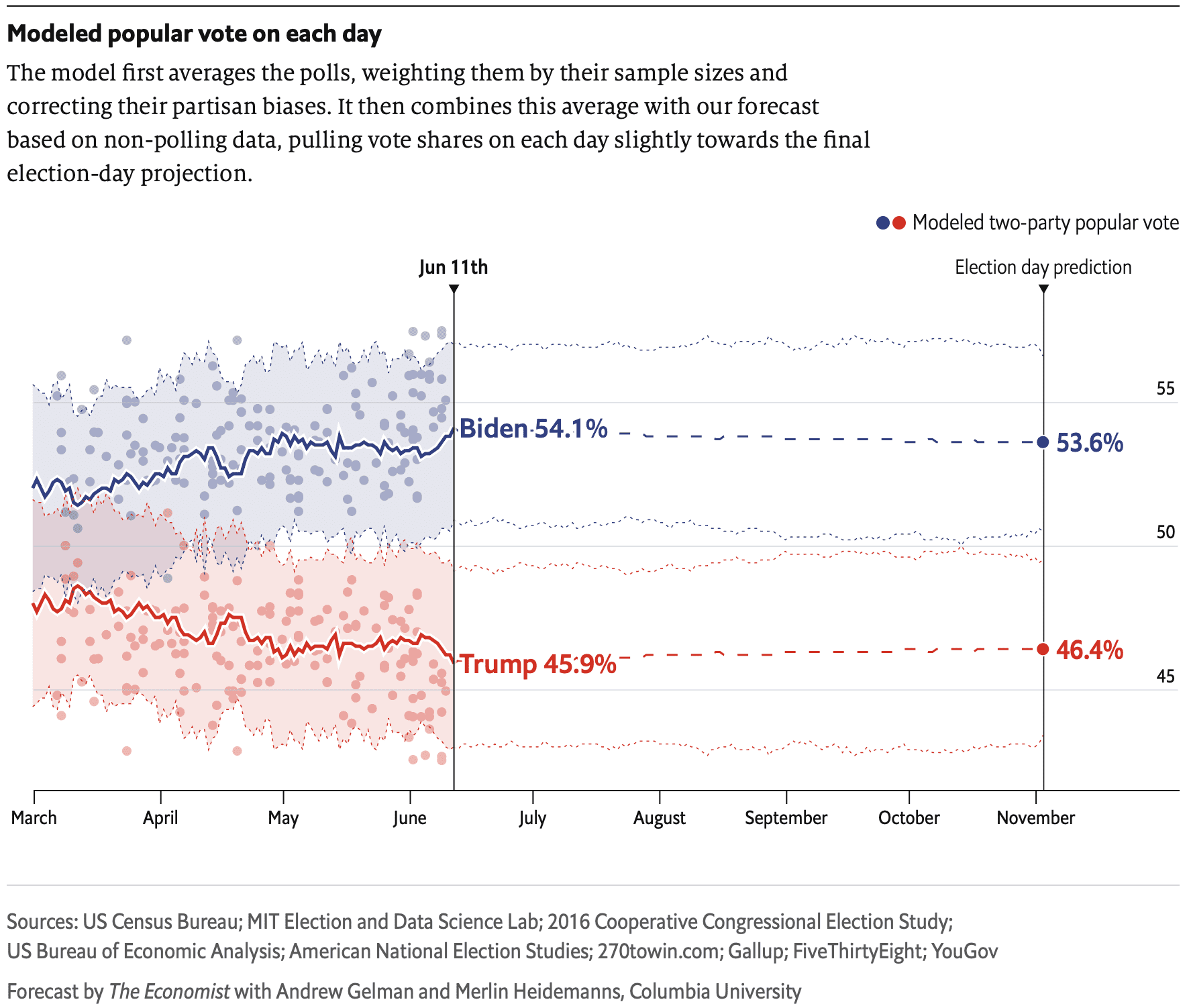}}
\caption{\em Our election forecast for {\em The Economist} on the day of its release in June 2020.  Before making this public forecast, we went through a series of checks of the data, model, and fitting procedure.  We made further changes to the model as the campaign went on.}
\label{fig:economist_1}
\end{figure}  

\subsubsection*{Future real-world goals.} 
In spring 2020 one of us collaborated with a data journalist at {\em The Economist} magazine to produce an ongoing state-by-state and national-level forecast of the upcoming U.S.\ presidential election \citep{heidemanns_gelman_morris_2020, gelman_hullman_wlezien_etal_2020}. The forecast automatically updated as new polls arose during the campaign.

\subsubsection*{Step 1 $\Rightarrow$ Training-data real-world goals.} 
A particular challenge of this problem is incorporating  diverse sources of data: national polls, state polls, economic statistics, and past state and national election results---as well as knowledge reflecting political science understanding of day-to-day changes in voting behavior and survey responses.

\paragraph{Trust.} By analogy to the microcredit example, trust could break down in this step if the data used in the model were deemed incomplete in the sense that there are other available data that could be used to noticeably improve the forecast.  For example, a forecast based entirely on national polls would have problems staying up to date with the latest state poll results.  Particularly relevant to the 2016 and 2020 elections were concerns about poll bias due to nonresponse; for instance, we might be concerned that fewer Republicans responded to polls, especially in key swing states.

\subsubsection*{Step 2 $\Rightarrow$ Mathematical goals} 
Constructing the model was more difficult than you might think. The probability distributions representing the underlying public opinion and survey data had to serve as a bridge connecting poll and economic data, political science understanding, and statistical structure (sampling error, nonsampling error, and variation of opinion over time in the 50 states).

Concretely, we started with an existing forecasting model in political science \citep{linzer2013dynamic}, which in turn was based on earlier models for poll aggregation and election forecasting.  The unique feature of our approach compared to other forecasts for the 2020 election was fitting a single model combining information from national polls, state polls, and economic and political fundamentals based on previous election outcomes.

Challenges in setting up this model included the following. We wanted to account for nonsampling error at both the state and national levels, including possible systematic biases favoring one party or the other; our error terms did not correct for biases, but they should allow the model to express forecasts with appropriate uncertainty. We aim to capture a time series of opinion changes during the campaign. We wanted to interpret the historical predictive power of economic fundamentals in the context of increasing political polarization. And we aimed to acknowledge unique features of the 2020 campaign.  A particular statistical challenge arose in modeling the correlations of polling errors and opinion trends across states:  there was nothing like the amount of data available to estimate a $50\times 50$ covariance matrix from any purely empirical procedure, so we knew that strong assumptions would be necessary.

\paragraph{Trust.} Trust could break down in this step if any of these strong assumptions or modeling choices could be replaced with other reasonable choices and lead us to substantively different conclusions.  Indeed, some of our initial modeling choices regarding between-state correlations were flawed, as we noticed after seeing some predictive intervals that seemed implausible.  Unfortunately, we only noticed some of these errors after the forecast went live, and we were forced to reboot our model and publicly explain our changes.

\subsubsection*{Step 3 $\Rightarrow$ Algorithms} 
As with the microcredit case study, we used HMC to approximate
and summarize the Bayesian posterior distribution on election outcomes and other posterior quantities of interest for political understanding.  The dataset and model were small enough and our time scale was gentle enough (requiring updates every day, not every hour or every minute) that there was no need to use a fast shortcut algorithm.

\paragraph{Trust.}
As discussed in the previous example, trust could break down if the approximation quality were poor but was not an issue here given the small size of the problem.

\subsubsection*{Step 4 $\Rightarrow$ Code} 
We used R and Stan for HMC; we also compiled and cleaned survey and election data.

\paragraph{Trust.}
Our general experience with Stan allowed us to trust its computation, so that the main concerns with the code were in the R scripts we wrote to prepare the data, set up covariance matrices, and postprocess the inferences.

\section{Establishing trust}
\label{sec:establishing_trust}

Above we have seen a number of ways that trust can break down across the taxonomy. But crucially there are also many ways for researchers to establish trust. We next place existing work into our taxonomy. We start from the bottom of \Cref{fig:trust} and work our way up. In particular, the italicized vertical connectors in \Cref{fig:trust} give terms related to trust for the corresponding step. The horizontal connectors in \Cref{fig:trust} show terms that describe matching between goals and outputs at this level. 

\subsection*{Algorithms and code} 
Trusting code to faithfully represent an algorithm (\emph{fidelity}) is sometimes considered prosaic and perhaps even outside the purview of the data scientist, especially when using standard software packages. Yet this step forms the bedrock of the analysis; if trust breaks down here, trust in the entire analysis will fail. And indeed bugs in data analyses have led to unsupported justification for treatment of patients in clinical trials \citep{baggerly2009deriving} and influential but unsupported economic policy advice \citep{herndon2014does,o2013forget}.

The name of the core issue here varies widely across the literature. In the present paper, we follow the usage of Conclusion 3-1 of \citet{national2019reproducibility}, labeling an analysis as \emph{reproducible} if identical results are achieved when the same data are analyzed again with the same code. Despite this definition seeming somewhat limited, many data analyses (including some of our own in busy research projects!) do not meet this standard. For instance, some data analyses do not provide code. Some do not even describe their algorithm in enough detail for the reader to produce equivalent code \citep{haibe2020transparency}. Even with full descriptions and code, checking reproducibility can be challenging and time-consuming.

Nonetheless, these observations suggest clear ways to build trust at this step: via open community engagement, code sharing, and data sharing. Even when using standard software packages, analysts typically write their own code wrappers to call these packages for the specific data and model at hand, and this new code needs to be checked. \citet{heil2021reproducibility} propose a gold standard where data analysts set up a single command that exactly reproduces the full analysis. Modern computational tools that manage models \citep{vartak2016modeldb} and aid reproducibility \citep{stodden2014implementing} can help researchers achieve the gold standard. It might be easy to overlook code that goes into data pre-processing, but this code is part of an analysis pipeline too; for instance, NASA failed to detect ozone reduction for much of the 1970s and 1980s (despite having appropriate sensors) due to an unexpected side effect of data pre-processing \citep{nasa2001ozone}.
\citet{gebru_morgenstern_vecchione_etal_2018} propose ``datasheets for datasets'' to fully detail data provenance, including any processing that may have gone into their creation.

During the election forecasting case study above, we opened up the process and results to a wide audience. Publication and daily updates in the Economist meant that thousands of readers would see each forecast.  Placing  data and code on GitHub allowed outsiders to download, run themselves, and explore places where something could be going wrong. Outside readers did indeed discover problems, which we were able to track down to bugs and conceptual problems with our model. And the sense of public responsibility motivated us to check carefully when forecasts did not look right.

\subsection*{Mathematical goals}
As we have seen in the case studies, a sufficiently complex Bayesian analysis typically aims to approximate summaries of the posterior.
\citet{geweke_2004}, \citet{cook_gelman_rubin_2006}, and \citet{talts_betancourt_simpson_etal_2018} present simulation-based methods for checking the \emph{accuracy} of a Bayesian approximation algorithm (via its implemented code). We and others have also developed methods for checking approximate computation on a particular dataset of interest; for example, \citet{yao_vehtari_simpson_etal_2018} and \citet{huggins2020validated} have presented methods for evaluating variational inference. These new approaches can help to identify mismatch between the mathematical goals and the joint implementation of the algorithm and code.

In the election forecasting case study, we simulated fake polling data and then saw how well the fitting procedure could recover the assumed parameters and underlying time series.  The check proceeded as follows: (1) we set hyperparameters of the chosen model to fixed, reasonable values; (2) we simulated time series of national and state public opinion using the generative time series including the between-state correlation matrix assumed under the model; (3) we simulated national and state polls at the same sample sizes and dates (in days before the election) as in the 2016 campaign, but with data set to the simulated underlying time series plus error, including both sampling and nonsampling error; and (4) we fit the model to these simulated data. It is no surprise that we should be able to roughly recover the hyperparameters and time series, but given the complexity of the model it is useful to check and also to get a sense of the precision of the recovery.

Practical checks at this level typically address algorithms and code together. Sometimes research will abstract the algorithm away from the code, for example by establishing \emph{theoretical guarantees} on how well a particular algorithm can be expected to achieve a goal such as accuracy or compute time. These guarantees can help establish trust. But they typically cannot be entirely relied upon to ensure accuracy: the exact set of assumptions needed for the guarantee may not hold in practice, or may not be possible to verify exactly in practice. 
A large amount of research concentrates at this level. Indeed, it is standard for any paper addressing Bayesian posterior approximation to come supported by theory. We likely see this abundance of research since the question of whether an algorithm achieves a precise mathematical goal has a relatively objective answer.

\subsection*{Training data and real-world goals}
To establish trust at this level, we want to ensure that our mathematical goals are meeting our real-world data analysis goals, at least on our training data.
Establishing trust here starts to become more subjective (and hence more challenging) relative to previous levels, since we must ask what it means to perform well on real-world goals, which are necessarily more abstract and ill-defined than the mathematical problems we reduce them to in Step 2. We emphasize this distinction with the color of connectors in \Cref{fig:trust}.

Perhaps due to the challenge of this subjectivity, we often do not see checks at this level in individual data analyses.
Nonetheless, there are lines of work that separately address these issues.
For instance, early work on Bayesian \emph{model checking} came from \citet{box_1980} and \citet{rubin_1984}; this research arose in an era when computations were relatively simple so the focus was on fit of model to data. In modern work, Bayesian model checking often assesses whether the joint model-algorithm-code mechanism does a good job of describing the available training data, sometimes with additional granularity \citep{gabry2019visualization}.

For instance, the election forecasting case study above used sense checks to assess the model-algorithm-code trifecta.
Some months elapsed from the start of our modeling process and \emph{The Economist}'s launch of the forecast.  This lag did not come from coding (as code is straightforward in Stan) or from running the model (the dataset including all polls was small enough that even without any real efforts at optimization the code ran in minutes) but rather from the steps of building and checking the model.  Once we had code that compiled and ran, we fit it to data from 2008, 2012, and 2016; these were earlier elections where we had a large supply of state and national polls so we could mimic the 2020 forecasting we planned to do.  Various early versions of our model produced results that did not make sense, for example with time trends seeming to vary too much from state to state, which implied that some variance parameter in our model was too large so that these time series were insufficiently constrained.  This checking was valuable for its own sake and also because, to do so, we needed to design graphs whose forms we would use when plotting models fit to the 2020 campaign data.

Before releasing our predictions to the public in early June 2020, we fit the model to the data available up to that point and checked that the inferences and our summary of them (see \Cref{fig:economist_1}) seemed reasonable, in the sense that they were consistent with our general understanding of the campaign and election.  Any model is only as good as the data it includes, so even if the fit has no statistical errors we would be concerned if it produced forecasts that were far off from our general beliefs.  We also examined the state-by-state forecasts from a competing model produced by the FiveThirtyEight team and found some implausible predictions, which we attributed to issues in how their model handled between-state correlations.

During the summer and fall of 2020, we continued to monitor our forecast.  In particular we were concerned that, months before election day, the estimated probability of Biden winning the electoral college was over 90\%, which did not seem to fully capture our uncertainties.  We looked carefully at our code and performed more simulations, re-fitting our model under alternative scenarios such as removing polls from one state at a time, and eventually we discovered some bugs in our code as well as other places where we were unsatisfied with the model, most notably in our expression of the between-state correlation matrix.  After a couple weeks of testing, we released an improved model along with a correction note on {\em The Economist} site.

Finally, we compared to the actual election outcomes. We found that our model performed well but not perfectly. The popular and electoral vote margins fell within our 90\% forecast intervals, with 48 out of 50 states predicted correctly. But the election was closer than our point forecast, with Biden performing consistently worse than predicted in almost every state.  Polls were off by about 2.5\% in two-party vote share across the country; our forecast did not anticipate this error, but it was included as a possibility in the correlated polling error model.  Including a correlated error term allowed the forecasts to acknowledge nonsampling error but is no substitute for predicting the direction of the bias.

These sense-making issues fundamentally relate to a form of stability: \emph{stability under different reasonable modeling choices}. 
\citet{yu2013stability}, \citet{yu_kumbier_2020}, and \citet{yu2020stability} have advocated for the importance of stability of many forms, including stability under model choice, in data analysis. 

We might be interested in stability across different representations or summaries of our data.
As an example, consider choosing among competing network embeddings; \citet{ward_huang_davison_etal_2021} examine this problem within the framework of \citet{yu_kumbier_2020}'s principles for veridical data science:  predicability, computability, and stability (PCS).
For some problems, we may be interested in embeddings that preserve particular features of the observed network or in embeddings that translate to better performance for some downstream task (e.g., link prediction), rather than in estimating a true underlying embedding \citep[or even mere features of that embedding, such as the manifold in which those embeddings lie, as in][]{smith_asta_calder_2019, lubold_chandrasekhar_mccormick_2020}.

We also often choose proxy measures to assess performance of a method, so we might wonder how stable our results are to these choices. We may choose proxies when questions in the social sciences, sciences, and engineering are difficult to operationalize directly---or when these questions might require substantial new modeling and inference development. For example, we might use a mean effect to understand the efficacy of microcredit since a mean is straightforward and standard to capture via a linear model---even though it might be more appropriate to judge the effect of microcredit by the proportion of people to whom it is beneficial.
As another example, many methods default to prediction with simple losses when the real-life goal or loss is difficult to quantify. For instance, 0-1 loss is a common choice of convenience in classification problems, but real-life classification loss is typically both unbalanced and difficult to quantify precisely.
Prediction is generally an easy (statistical or mathematical) problem precisely because it is defined in terms of a clear ground truth that we can compare our predictions to.
Even problems that are naturally framed as prediction, like medical diagnosis, may not be appropriately translated into a concrete optimization problem; e.g., consider a model for detecting cancerous skin legions that inadvertently trains on surgical ink markings \citep{winkler_fink_toberer_etal_2019}.

These proxy measures of performance form part of a larger phenomenon in scientific research:  hard questions often get translated into questions that are easier to answer but possibly substantively further from the question we want to ask.  At its core, this critique is related to the question of how problem formulation can impact scientific outcomes \citep{passi_barocas_2019}.  More broadly, it is important to consider whether we care about prediction or explanation \citep{shmueli_2010}: a particular analyst's real-world goals might be best addressed by optimizing a black box that makes highly accurate predictions, or by constructing a more \emph{interpretable} model with strong explanatory power, or perhaps by doing something in between.

To test stability to these various choices in practice, \citet{silberzahn_uhlmann_martin_etal_2018} documents variations among twenty-nine distinct research teams' analyses of a shared real-world goal, equipped with shared training data: whether soccer referees are more likely to give red cards to players with dark skin tones.
Variations in adopted analytic strategies ranged in breadth from differences in which predictors are included, to how dependent observations are (or are not) adjusted for, to statistical modeling assumptions (e.g., linear or logistic regression). 

Convening multiple research teams for a problem is not always feasible, at least not at first. So we might wish to have more automated checks usable by a single team. For instance, in \citet{smith_zheng_gelman_2020}, we construct prediction scores as a measure of agreement between data-generating mechanisms (e.g., preregistration or pilot data and realized experimental data); we show that modifications to the predictive model (used to construct the predictions) adjust the lens through which the prediction scores can pick up on varying types of differences between the underlying data generating mechanisms. This check goes beyond the mere reproduction of numerical outputs, but more broadly accounts for how the real world interacts with the inferential chain:  how the real world goals are interpreted and formulated as mathematical goals.

Another body of literature looks at quantifying \emph{robustness} of the conclusions of a data analysis \emph{to likelihood and prior choice} in a Bayesian analysis \citep{berger2000bayesian}, with tools to interface with MCMC \citep{gustafson2000local} and variational methods \citep{giordano2018covariances,giordano2021evaluating}. A tricky aspect of operationalizing this work is deciding how to mathematically capture a range of reasonable models; the easiest option is to vary a hyperparameter of a model (that may control the model parametrically or nonparametrically) over some range that the analyst deems reasonable. Other options include visualizations to aid analysts and domain experts in making sense checks \citep{gabry2019visualization,stephenson2021measuring}.

\subsection*{Future real-world goals}

Again, building on Conclusion 3-1 in \cite{national2019reproducibility}, we say that an analysis is \textit{replicable} if similar results are achieved when the study is repeated with fresh data.\footnote{A replicable study does not strictly require reproducibility (or matching of goals and outputs at other levels of \Cref{fig:trust}). But it is hard to imagine how a non-reproducible analysis could be successfully replicated.} We can still speak of \emph{stability and robustness}, but now \emph{to changing data} rather than changing model choices.
For instance, ultimately we wish to apply the conclusions of our microcredit case study to decide whether to pursue microloans in new places in the future---or the same places but at necessarily different (future) times. If we concluded that microcredit alleviated poverty based on our original data, can we continue to trust these conclusions?

As in the previous case, the best option for establishing replicability is to run the experiment many times in many different conditions. 
\citet{OpenScienceCollaboration_2015} brought replicability to the forefront of the scientific discussion via an attempt to replicate one hundred high-impact psychological studies.
But these re-runs can be costly, and they do not alleviate the need to make, e.g., policy decisions using past analyses. To that end, there is a need for automated tools that can assess sensitivity and robustness to data changes that might reasonably reflect changes we expect to see before applying policy.

Classical tools such as cross validation and the bootstrap can provide a notion of sensitivity by re-running a data analysis many times. But they reflect an implicit assumption that the data-generating distribution of new data is fundamentally the same as that of existing data. This assumption is often at least somewhat inappropriate. Recall that we hope to capture and apply universal truths from a data analysis. We therefore might not apply decisions from our data analysis to a wildly different context. But we still might expect, and wish to anticipate, small but substantive changes across data sets of interest. 
\citet{damour_heller_moldovan_etal_2020} have found that many conclusions of AI analyses do not hold up when used in real life despite being trained and tested in standard machine learning pipelines.
In our own work, we reasonably expect regional differences in the effects of microcredit.

\citet{damour_heller_moldovan_etal_2020} suggest stress-testing data analyses to explore performance in practically relevant dimensions and to attempt to identify potential  inductive  biases.
One option that we have developed and applied to our microcredit case study is assessing sensitivity of an analysis to dropping a small amount of data \citep{broderick_giordano_meager_2020}. If an analysis is driven by a small proportion of its data, we might not expect it to generalize well to other scenarios. In theory, this sensitivity can be checked directly by re-running the data analysis with every small proportion of data removed; in practice, this naive approach is astronomically costly on even small data sets. So we develop an approximation that can be checked directly: when it detects sensitivity, it returns the problematic small subset of data. So the analyst can drop this small data subset and re-run the analysis at the cost of just one additional data analysis.

We applied exactly this dropping-data check to our analysis of microcredit in \citep{meager2020aggregating}. We hoped that various features of the analysis would decrease sensitivity: hierarchical Bayesian sharing of power across datasets, regularization from priors, and a tailored likelihood meant to more carefully reflect the data-generating process. But we still found that the average effects were sensitive to dropping less than 1\% of the data---but the estimated variance in treatment effects across studies was more robust.

\section{Discussion}
Perfect accuracy is not a requirement for trust.
Indeed, we are able to trust many of the systems we interact with in our day-to-day lives despite their uncertain predictions: weather predictions are not always correct, and yet they are still useful; many people are injured or die in car crashes, and yet we rely on cars to travel from place to place; we do not expect doctors to be able to comprehensively restore us to our previous health, and yet we trust our doctors.
Even for probability models, our thresholds for acceptable accuracy may depend on the setting or type of model.
For example, much of classical statistics developed out of agricultural settings.
As a result, many statistical models are well-aligned with the physical sciences and essentially act as deterministic or mechanistic process models with noise.  In these models, the error is often treated like a nuisance parameter and included because we wouldn't be able to estimate the model otherwise.
Other statistical models are more descriptive and are fundamentally models of variability, such as in economics or the social sciences.
A third class of models is even more disconnected from observable physical processes;  these discovery or exploratory models are specifically designed to capture fuzziness.  Examples include clustering models, topic models, and other unsupervised learning approaches.  We naturally expect different levels of accuracy for models in each of these classes and, especially for discovery or exploratory models, defining accuracy can be more difficult.

More generally, we emphasize that trust itself is not binary. We have here laid out a variety of points in an analysis where trust can fail or be increased, but even at each point trust lies on a spectrum. \citet{heil2021reproducibility} note that, even at the level of code reproducibility, there lies a spectrum of trust: from not providing code or details to providing them but without easy documentation or use all the way to providing well-documented code that duplicates an existing analysis with a single command. We analogously argue in \cite{smith_zheng_gelman_2020} for treating trust as a spectrum, rather than a binary, at other levels of our present taxonomy.
In almost all settings, there will be some variation (due to different operationalizations of the real world goals, or to stochastic elements in the algorithm, or even to random number generation in the code), even if only slightly.  Instead, we should focus on quantifying {\em how much} variation is present. 
In \cite{giordano2018covariances}, we make the distinction between sensitivity and robustness as follows: sensitivity measures the (continuous or near-continuous) degree of change in some reported value due to, e.g., changing the model. That is, sensitivity is a well-defined continuous quantity. But robustness is a subjective judgment call based on sensitivity; we say an analysis is non-robust if a particular observed change is deemed important in that a change in reported value affects a particular actionable decision. We see a version of this in our check for dropping data \citep{broderick_giordano_meager_2020}. We assess sensitivity by checking the following: for different sizes of dropped data subsets, what is the biggest change we expect to see in our reported quantity of interest. Ultimately an analyst must  decide if robustness is a concern by asking whether the change in data or model required to see a substantively different conclusion is too small.

\clearpage
\bibliography{compare}

\begin{thebibliography}{52}
\newcommand{\enquote}[1]{``#1''}
\expandafter\ifx\csname natexlab\endcsname\relax\def\natexlab#1{#1}\fi
\expandafter\ifx\csname url\endcsname\relax
  \def\url#1{{\tt #1}}\fi
\expandafter\ifx\csname urlprefix\endcsname\relax\def\urlprefix{URL }\fi

\bibitem[{Baggerly and Coombes(2009)}]{baggerly2009deriving}
Baggerly, K.~A. and Coombes, K.~R.
\newblock \enquote{Deriving chemosensitivity from cell lines: Forensic
  bioinformatics and reproducible research in high-throughput biology.}
\newblock {\em Annals of Applied Statistics\/}, 1309--1334 (2009).

\bibitem[{Berger et~al.(2000)Berger, Insua, and Ruggeri}]{berger2000bayesian}
Berger, J.~O., Insua, D.~R., and Ruggeri, F.
\newblock \enquote{Bayesian robustness.}
\newblock In {\em Robust Bayesian Analysis\/}, 1--32. Springer (2000).

\bibitem[{Blei et~al.(2017)Blei, Kucukelbir, and
  McAuliffe}]{blei2017variational}
Blei, D.~M., Kucukelbir, A., and McAuliffe, J.~D.
\newblock \enquote{Variational inference: A review for statisticians.}
\newblock {\em Journal of the American Statistical Association\/},
  112(518):859--877 (2017).

\bibitem[{Box(1980)}]{box_1980}
Box, G. E.~P.
\newblock \enquote{Sampling and Bayes' inference in scientific modelling and
  robustness.}
\newblock {\em Journal of the Royal Statistical Society A\/}, 143(4):383--404
  (1980).

\bibitem[{Broderick et~al.(2020)Broderick, Giordano, and
  Meager}]{broderick_giordano_meager_2020}
Broderick, T., Giordano, R., and Meager, R.
\newblock \enquote{An automatic finite-sample robustness metric: Can dropping a
  little data change conclusions?}
\newblock {\em arXiv preprint arXiv:2011.14999\/} (2020).

\bibitem[{Cook et~al.(2006)Cook, Gelman, and Rubin}]{cook_gelman_rubin_2006}
Cook, S.~R., Gelman, A., and Rubin, D.~B.
\newblock \enquote{Validation of software for Bayesian models using posterior
  quantiles.}
\newblock {\em Journal of Computational and Graphical Statistics\/},
  15(3):675--692 (2006).

\bibitem[{D'Amour et~al.(2020)D'Amour, Heller, Moldovan, Adlam, Alipanahi,
  Beutel, Chen, Deaton, Eisenstein, Hoffman, Hormozdiari, Houlsby, Hou, Jerfel,
  Karthikesalingam, Lucic, Ma, McLean, Mincu, Mitani, Montanari, Nado,
  Natarajan, Nielson, Osborne, Raman, Ramasamy, Sayres, Schrouff, Seneviratne,
  Sequeira, Suresh, Veitch, Vladymyrov, Wang, Webster, Yadlowsky, Yun, Zhai,
  and Sculley}]{damour_heller_moldovan_etal_2020}
D'Amour, A., Heller, K.~A., Moldovan, D., Adlam, B., Alipanahi, B., Beutel, A.,
  Chen, C., Deaton, J., Eisenstein, J., Hoffman, M.~D., Hormozdiari, F.,
  Houlsby, N., Hou, S., Jerfel, G., Karthikesalingam, A., Lucic, M., Ma, Y.,
  McLean, C.~Y., Mincu, D., Mitani, A., Montanari, A., Nado, Z., Natarajan, V.,
  Nielson, C., Osborne, T.~F., Raman, R., Ramasamy, K., Sayres, R., Schrouff,
  J., Seneviratne, M., Sequeira, S., Suresh, H., Veitch, V., Vladymyrov, M.,
  Wang, X., Webster, K., Yadlowsky, S., Yun, T., Zhai, X., and Sculley, D.
\newblock \enquote{Underspecification presents challenges for credibility in
  modern machine learning.}
\newblock {\em arXiv preprint arXiv:2011.03395\/} (2020).

\bibitem[{Fischer et~al.(2020)Fischer, Morris, van Doremalen, Sarchette,
  Matson, Bushmaker, Yinda, Seifert, Gamble, Williamson, Judson, de~Wit,
  Lloyd-Smith, and Munster}]{fischer2020assessment}
Fischer, R.~J., Morris, D.~H., van Doremalen, N., Sarchette, S., Matson, M.~J.,
  Bushmaker, T., Yinda, C.~K., Seifert, S.~N., Gamble, A., Williamson, B.~N.,
  Judson, S.~D., de~Wit, E., Lloyd-Smith, J.~O., and Munster, V.~J.
\newblock \enquote{Assessment of N95 respirator decontamination and re-use for
  SARS-CoV-2.}
\newblock {\em MedRxiv\/} (2020).

\bibitem[{Flaxman et~al.(2020)Flaxman, Mishra, Gandy, Unwin, Coupland, Mellan,
  Zhu, Berah, Eaton, Guzman, Schmit, Callizo, Ainslie, Baguelin, Blake,
  Boonyasiri, Boyd, Cattarino, Ciavarella, Cooper, Cucunub\'{a},
  Cuomo-Dannenburg, Dighe, Djaafara, Dorigatti, van Elsland, FitzJohn, Fu,
  Gaythorpe, Geidelberg, Grassly, Green, Hallett, Hamlet, Hinsley, Jeffrey,
  Jorgensen, Knock, Laydon, Nedjati-Gilani, Nouvellet, Parag, Siveroni,
  Thompson, Verity, Volz, Walker, Walters, Wang, Wang, Watson, Whittaker,
  Winskill, Xi, Ghani, Donnelly, Riley, Okell, Vollmer, Ferguson, and
  Bhatt}]{flaxman2020report}
Flaxman, S., Mishra, S., Gandy, A., Unwin, H. J.~T., Coupland, H., Mellan,
  T.~A., Zhu, H., Berah, T., Eaton, J.~W., Guzman, P. N.~P., Schmit, N.,
  Callizo, L., Ainslie, K. E.~C., Baguelin, M., Blake, I., Boonyasiri, A.,
  Boyd, O., Cattarino, L., Ciavarella, C., Cooper, L., Cucunub\'{a}, Z.,
  Cuomo-Dannenburg, G., Dighe, A., Djaafara, B., Dorigatti, I., van Elsland,
  S., FitzJohn, R., Fu, H., Gaythorpe, K., Geidelberg, L., Grassly, N., Green,
  W., Hallett, T., Hamlet, A., Hinsley, W., Jeffrey, B., Jorgensen, D., Knock,
  E., Laydon, D., Nedjati-Gilani, G., Nouvellet, P., Parag, K., Siveroni, I.,
  Thompson, H., Verity, R., Volz, E., Walker, P.~G., Walters, C., Wang, H.,
  Wang, Y., Watson, O., Whittaker, C., Winskill, P., Xi, X., Ghani, A.,
  Donnelly, C.~A., Riley, S., Okell, L.~C., Vollmer, M. A.~C., Ferguson, N.~M.,
  and Bhatt, S.
\newblock \enquote{Report 13: Estimating the number of infections and the
  impact of non-pharmaceutical interventions on COVID-19 in 11 European
  countries.} (2020).

\bibitem[{Gabry et~al.(2019)Gabry, Simpson, Vehtari, Betancourt, and
  Gelman}]{gabry2019visualization}
Gabry, J., Simpson, D., Vehtari, A., Betancourt, M., and Gelman, A.
\newblock \enquote{Visualization in Bayesian workflow.}
\newblock {\em Journal of the Royal Statistical Society A\/}, 182(2):389--402
  (2019).

\bibitem[{Gebru et~al.(2018)Gebru, Morgenstern, Vecchione, Vaughan, Wallach,
  Daum{\'e}~III, and Crawford}]{gebru_morgenstern_vecchione_etal_2018}
Gebru, T., Morgenstern, J., Vecchione, B., Vaughan, J.~W., Wallach, H.,
  Daum{\'e}~III, H., and Crawford, K.
\newblock \enquote{Datasheets for datasets.}
\newblock {\em arXiv preprint arXiv:1803.09010\/} (2018).

\bibitem[{Gelman et~al.(2020)Gelman, Hullman, Wlezien, and
  Morris}]{gelman_hullman_wlezien_etal_2020}
Gelman, A., Hullman, J., Wlezien, C., and Morris, G.~E.
\newblock \enquote{Information, incentives, and goals in election forecasts.}
\newblock {\em Judgment and Decision Making\/}, 15(5):863 (2020).

\bibitem[{Geweke(2004)}]{geweke_2004}
Geweke, J.
\newblock \enquote{Getting it right: Joint distribution tests of posterior
  simulators.}
\newblock {\em Journal of the American Statistical Association\/},
  99(467):799--804 (2004).

\bibitem[{Giordano et~al.(2018)Giordano, Broderick, and
  Jordan}]{giordano2018covariances}
Giordano, R., Broderick, T., and Jordan, M.~I.
\newblock \enquote{Covariances, robustness and variational Bayes.}
\newblock {\em Journal of Machine Learning Research\/}, 19(51) (2018).

\bibitem[{Giordano et~al.(2016)Giordano, Broderick, Meager, Huggins, and
  Jordan}]{giordano2016fast}
Giordano, R., Broderick, T., Meager, R., Huggins, J., and Jordan, M.
\newblock \enquote{Fast robustness quantification with variational Bayes.}
\newblock In {\em ICML 2016 Workshop on \#Data4Good: Machine Learning in Social
  Good Applications\/} (2016).

\bibitem[{Giordano et~al.(2021)Giordano, Liu, Jordan, and
  Broderick}]{giordano2021evaluating}
Giordano, R., Liu, R., Jordan, M.~I., and Broderick, T.
\newblock \enquote{Evaluating sensitivity to the stick-breaking prior in
  Bayesian nonparametrics.}
\newblock {\em arXiv preprint arXiv:1810.06587\/} (2021).

\bibitem[{Giordano et~al.(2015)Giordano, Broderick, and
  Jordan}]{giordano2015linear}
Giordano, R.~J., Broderick, T., and Jordan, M.~I.
\newblock \enquote{Linear response methods for accurate covariance estimates
  from mean field variational Bayes.}
\newblock In {\em Neural Information Processing Systems\/} (2015).

\bibitem[{Gustafson(2000)}]{gustafson2000local}
Gustafson, P.
\newblock \enquote{Local robustness in Bayesian analysis.}
\newblock In {\em Robust Bayesian Analysis\/}, 71--88. Springer (2000).

\bibitem[{Haibe-Kains et~al.(2020)Haibe-Kains, Adam, Hosny, Khodakarami,
  Waldron, Wang, McIntosh, Goldenberg, Kundaje, Greene, Broderick, Hoffman,
  Leek, Korthauer, Huber, Brazma, Pineau, Tibshirani, Hastie, Ioannidis,
  Quackenbush, and Aerts}]{haibe2020transparency}
Haibe-Kains, B., Adam, G.~A., Hosny, A., Khodakarami, F., Waldron, L., Wang,
  B., McIntosh, C., Goldenberg, A., Kundaje, A., Greene, C.~S., Broderick, T.,
  Hoffman, M.~M., Leek, J.~T., Korthauer, K., Huber, W., Brazma, A., Pineau,
  J., Tibshirani, R., Hastie, T., Ioannidis, J. P.~A., Quackenbush, J., and
  Aerts, H. J. W.~L.
\newblock \enquote{Transparency and reproducibility in artificial
  intelligence.}
\newblock {\em Nature\/}, 586(7829):E14--E16 (2020).

\bibitem[{Heidemanns et~al.(2020)Heidemanns, Gelman, and
  Morris}]{heidemanns_gelman_morris_2020}
Heidemanns, M., Gelman, A., and Morris, G.~E.
\newblock \enquote{An updated dynamic Bayesian forecasting model for the 2020
  election.}
\newblock {\em Harvard Data Science Review\/}, 2 (2020).

\bibitem[{Heil et~al.(2021)Heil, Hoffman, Markowetz, Lee, Greene, and
  Hicks}]{heil2021reproducibility}
Heil, B.~J., Hoffman, M.~M., Markowetz, F., Lee, S.-I., Greene, C.~S., and
  Hicks, S.~C.
\newblock \enquote{Reproducibility standards for machine learning in the life
  sciences.}
\newblock {\em Nature Methods\/}, 18(10):1132--1135 (2021).

\bibitem[{Herndon et~al.(2014)Herndon, Ash, and Pollin}]{herndon2014does}
Herndon, T., Ash, M., and Pollin, R.
\newblock \enquote{Does high public debt consistently stifle economic growth? A
  critique of Reinhart and Rogoff.}
\newblock {\em Cambridge Journal of Economics\/}, 38(2):257--279 (2014).

\bibitem[{Huggins et~al.(2020)Huggins, Kasprzak, Campbell, and
  Broderick}]{huggins2020validated}
Huggins, J., Kasprzak, M., Campbell, T., and Broderick, T.
\newblock \enquote{Validated variational inference via practical posterior
  error bounds.}
\newblock In {\em International Conference on Artificial Intelligence and
  Statistics\/}, 1792--1802. PMLR (2020).

\bibitem[{Linzer(2013)}]{linzer2013dynamic}
Linzer, D.~A.
\newblock \enquote{Dynamic Bayesian forecasting of presidential elections in
  the states.}
\newblock {\em Journal of the American Statistical Association\/},
  108(501):124--134 (2013).

\bibitem[{Lubold et~al.(2020)Lubold, Chandrasekhar, and
  McCormick}]{lubold_chandrasekhar_mccormick_2020}
Lubold, S., Chandrasekhar, A.~G., and McCormick, T.~H.
\newblock \enquote{Identifying the latent space geometry of network models
  through analysis of curvature.}
\newblock Technical report, National Bureau of Economic Research (2020).

\bibitem[{Meager(2019)}]{meager_2019}
Meager, R.
\newblock \enquote{Understanding the average impact of microcredit expansions:
  A Bayesian hierarchical analysis of seven randomized experiments.}
\newblock {\em American Economic Journal: Applied Economics\/}, 11(1):57--91
  (2019).

\bibitem[{Meager(2020)}]{meager2020aggregating}
---.
\newblock \enquote{Aggregating distributional treatment effects: A Bayesian
  hierarchical analysis of the microcredit literature.}
\newblock {\em Working paper\/} (2020).

\bibitem[{{NASA Earth Observatory}(2001)}]{nasa2001ozone}
{NASA Earth Observatory}.
\newblock \enquote{Research satellites for atmospheric sciences, 1978--present:
  Serendipity and stratospheric ozone.}
\newblock
  \url{https://earthobservatory.nasa.gov/features/RemoteSensingAtmosphere/remote_sensing5.php}
  (2001).

\bibitem[{{National Academies of Sciences, Engineering, and
  Medicine}(2019)}]{national2019reproducibility}
{National Academies of Sciences, Engineering, and Medicine}.
\newblock {\em Reproducibility and Replicability in Science\/}.
\newblock National Academies Press (2019).

\bibitem[{Neal(2011)}]{neal2011mcmc}
Neal, R.~M.
\newblock \enquote{MCMC using Hamiltonian dynamics.}
\newblock {\em Handbook of Markov chain Monte Carlo\/}, 2(11):2 (2011).

\bibitem[{O'Brien(2013)}]{o2013forget}
O'Brien, M.
\newblock \enquote{Forget Excel: This was Reinhart and Rogoff's biggest
  mistake.}
\newblock {\em The Atlantic\/}, 18 (2013).

\bibitem[{{Open Science Collaboration}(2015)}]{OpenScienceCollaboration_2015}
{Open Science Collaboration}.
\newblock \enquote{Estimating the reproducibility of psychological science.}
\newblock {\em Science\/}, 349(6251) (2015).

\bibitem[{Passi and Barocas(2019)}]{passi_barocas_2019}
Passi, S. and Barocas, S.
\newblock \enquote{Problem formulation and fairness.}
\newblock In {\em Proceedings of the Conference on Fairness, Accountability,
  and Transparency\/}, 39--48 (2019).

\bibitem[{Reinhart and Rogoff(2010)}]{reinhart2010growth}
Reinhart, C.~M. and Rogoff, K.~S.
\newblock \enquote{Growth in a time of debt.}
\newblock {\em American economic review\/}, 100(2):573--78 (2010).

\bibitem[{Rubin(1984)}]{rubin_1984}
Rubin, D.~B.
\newblock \enquote{Bayesianly justifiable and relevant frequency calculations
  for the applied statistician.}
\newblock {\em Annals of Statistics\/}, 1151--1172 (1984).

\bibitem[{Schwartz et~al.(2021)Schwartz, Down, Jonas, and
  Tabassi}]{schwartz_down_jonas_etal_2021}
Schwartz, R., Down, L., Jonas, A., and Tabassi, E.
\newblock \enquote{A proposal for identifying and managing bias within
  artificial intelligence.}
\newblock {\em NIST Special Publication\/}, 1270 (2021).

\bibitem[{Shmueli(2010)}]{shmueli_2010}
Shmueli, G.
\newblock \enquote{To explain or to predict?}
\newblock {\em Statistical Science\/}, 25(3):289--310 (2010).

\bibitem[{Silberzahn et~al.(2018)Silberzahn, Uhlmann, Martin, Anselmi, Aust,
  Awtrey, Bahn{\'\i}k, Bai, Bannard, Bonnier
  et~al.}]{silberzahn_uhlmann_martin_etal_2018}
Silberzahn, R., Uhlmann, E.~L., Martin, D.~P., Anselmi, P., Aust, F., Awtrey,
  E., Bahn{\'\i}k, {\v{S}}., Bai, F., Bannard, C., Bonnier, E., et~al.
\newblock \enquote{Many analysts, one data set: Making transparent how
  variations in analytic choices affect results.}
\newblock {\em Advances in Methods and Practices in Psychological Science\/},
  1(3):337--356 (2018).

\bibitem[{Smith et~al.(2019)Smith, Asta, and Calder}]{smith_asta_calder_2019}
Smith, A.~L., Asta, D.~M., and Calder, C.~A.
\newblock \enquote{The geometry of continuous latent space models for network
  data.}
\newblock {\em Statistical Science\/}, 34(3):428 (2019).

\bibitem[{Smith et~al.(2020)Smith, Zheng, and Gelman}]{smith_zheng_gelman_2020}
Smith, A.~L., Zheng, T., and Gelman, A.
\newblock \enquote{Prediction scoring for measuring the replicability of
  data-driven discoveries.}
\newblock {\em Columbia Academic Commons doi:10.7916/D82F95D0\/} (2020).

\bibitem[{{Stan Development Team}(2021)}]{stan}
{Stan Development Team}.
\newblock \enquote{Stan Modeling Language Users Guide and Reference Manual.}
  (2021).
\newblock Version 2.27.

\bibitem[{Stephenson et~al.(2021)Stephenson, Ghosh, Nguyen, Yurochkin,
  Deshpande, and Broderick}]{stephenson2021measuring}
Stephenson, W.~T., Ghosh, S., Nguyen, T.~D., Yurochkin, M., Deshpande, S.~K.,
  and Broderick, T.
\newblock \enquote{Measuring the sensitivity of Gaussian processes to kernel
  choice.}
\newblock {\em arXiv preprint arXiv:2106.06510\/} (2021).

\bibitem[{Stodden et~al.(2014)Stodden, Leisch, and
  Peng}]{stodden2014implementing}
Stodden, V., Leisch, F., and Peng, R.~D.
\newblock {\em Implementing Reproducible Research\/}.
\newblock CRC Press (2014).

\bibitem[{Talts et~al.(2018)Talts, Betancourt, Simpson, Vehtari, and
  Gelman}]{talts_betancourt_simpson_etal_2018}
Talts, S., Betancourt, M., Simpson, D., Vehtari, A., and Gelman, A.
\newblock \enquote{Validating {Bayesian} inference algorithms with
  simulation-based calibration.}
\newblock {\em arXiv preprint arXiv:1804.06788\/} (2018).

\bibitem[{Van~Doremalen et~al.(2020)Van~Doremalen, Bushmaker, Morris, Holbrook,
  Gamble, Williamson, Tamin, Harcourt, Thornburg, Gerber, Lloyd-Smith, de~Wit,
  and Munster}]{van2020aerosol}
Van~Doremalen, N., Bushmaker, T., Morris, D.~H., Holbrook, M.~G., Gamble, A.,
  Williamson, B.~N., Tamin, A., Harcourt, J.~L., Thornburg, N.~J., Gerber,
  S.~I., Lloyd-Smith, J.~O., de~Wit, E., and Munster, V.~J.
\newblock \enquote{Aerosol and surface stability of SARS-CoV-2 as compared with
  SARS-CoV-1.}
\newblock {\em New England Journal of Medicine\/}, 382(16):1564--1567 (2020).

\bibitem[{Vartak et~al.(2016)Vartak, Subramanyam, Lee, Viswanathan, Husnoo,
  Madden, and Zaharia}]{vartak2016modeldb}
Vartak, M., Subramanyam, H., Lee, W.-E., Viswanathan, S., Husnoo, S., Madden,
  S., and Zaharia, M.
\newblock \enquote{ModelDB: a system for machine learning model management.}
\newblock In {\em Proceedings of the Workshop on Human-In-the-Loop Data
  Analytics\/}, 1--3 (2016).

\bibitem[{Ward et~al.(2021)Ward, Huang, Davison, and
  Zheng}]{ward_huang_davison_etal_2021}
Ward, O.~G., Huang, Z., Davison, A., and Zheng, T.
\newblock \enquote{Next waves in veridical network embedding.}
\newblock {\em Statistical Analysis and Data Mining: The ASA Data Science
  Journal\/}, 14(1):5--17 (2021).

\bibitem[{Winkler et~al.(2019)Winkler, Fink, Toberer, Enk, Deinlein,
  Hofmann-Wellenhof, Thomas, Lallas, Blum, Stolz, and
  Haenssle}]{winkler_fink_toberer_etal_2019}
Winkler, J.~K., Fink, C., Toberer, F., Enk, A., Deinlein, T.,
  Hofmann-Wellenhof, R., Thomas, L., Lallas, A., Blum, A., Stolz, W., and
  Haenssle, H.~A.
\newblock \enquote{Association between surgical skin markings in dermoscopic
  images and diagnostic performance of a deep learning convolutional neural
  network for melanoma recognition.}
\newblock {\em JAMA Dermatology\/}, 155(10):1135--1141 (2019).

\bibitem[{Yao et~al.(2018)Yao, Vehtari, Simpson, and
  Gelman}]{yao_vehtari_simpson_etal_2018}
Yao, Y., Vehtari, A., Simpson, D., and Gelman, A.
\newblock \enquote{Yes, but did it work?: Evaluating variational inference.}
\newblock In {\em International Conference on Machine Learning\/}, 5581--5590.
  PMLR (2018).

\bibitem[{Yu(2013)}]{yu2013stability}
Yu, B.
\newblock \enquote{Stability.}
\newblock {\em Bernoulli\/}, 19(4):1484--1500 (2013).

\bibitem[{Yu(2020)}]{yu2020stability}
---.
\newblock \enquote{Stability expanded, in reality.}
\newblock {\em Harvard Data Science Review\/} (2020).

\bibitem[{Yu and Kumbier(2020)}]{yu_kumbier_2020}
Yu, B. and Kumbier, K.
\newblock \enquote{Veridical data science.}
\newblock {\em Proceedings of the National Academy of Sciences\/},
  117(8):3920--3929 (2020).

\end{thebibliography}

\end{document}